# Redes Generativas Adversarias (GAN) Fundamentos Teóricos y Aplicaciones

Survey

**Jordi de la Torre**[*]
Ph.D. in Computer Science (ML/AI)
Universitat Oberta de Catalunya
Barcelona, ES
jordi.delatorre@gmail.com

February 18, 2023

## Abstract

Las redes adversarias generativas (GANs) son un método basado en el entrenamiento de dos redes neuronales, una denominada generadora y otra discriminadora, compitiendo entre sí para generar nuevas instancias que se asemejen a las de la distribución de probabilidad de los datos de entrenamiento. Las GANs tienen una amplia gama de aplicaciones en campos como la visión por computadora, la segmentación semántica, la síntesis de series temporales, la edición de imagen, el procesamiento del lenguaje natural y la generación de imagen a partir de texto, entre otros. Los modelos generativos modelizan la distribución de probabilidad de un conjunto de datos, pero en lugar de proporcionar un valor de probabilidad, generan nuevas instancias cercanas a la distribución original. Las GANs utilizan un esquema de aprendizaje que permite codificar los atributos definitorios de la distribución de probabilidad en una red neuronal, lo que permite generar instancias que se asemejen a la distribución de probabilidad original. En este artículo se presentan los fundamentos teóricos de este tipo de redes así como los esquemas básicos de la arquitectura y algunas de sus aplicaciones. Este artículo está en español para facilitar la llegada de este conocimiento científico a la comunidad hispanohablante.

## Abstract

Generative adversarial networks (GANs) are a method based on the training of two neural networks, one called generator and the other discriminator, competing with each other to generate new instances that resemble those of the probability distribution of the training data. GANs have a wide range of applications in fields such as computer vision, semantic segmentation, time series synthesis, image editing, natural language processing, and image generation from text, among others. Generative models model the probability distribution of a data set, but instead of providing a probability value, they generate new instances that are close to the original distribution. GANs use a learning scheme that allows the defining attributes of the probability distribution to be encoded in a neural network, allowing instances to be generated that resemble the original probability distribution. This article presents the theoretical foundations of this type of network as well as the basic architecture schemes and some of its applications. This article is in Spanish to facilitate the arrival of this scientific knowledge to the Spanish-speaking community.

***Keywords*** redes generativas · GAN · entrenamiento adversario · inteligencia artificial · machine learning

---

[*]mailto:jordi.delatorre@gmail.com web:jorditg.github.io

# 1 Introducción

Las redes antagónicas generativas o redes adversarias generativas (GANs) ([1], [2], [3]) son un método para la optimización competitivo entre dos redes neuronales, una llamada generadora y otra discriminadora, con el objetivo de conseguir generar nuevas instancias idealmente indistinguibles a las pertenecientes a la distribución de probabilidad de la que derivan los datos de entrenamiento.

El fundamento teórico general del que derivan, permite su utilización para la generación de cualquier tipo de datos, habiéndose demostrado efectiva en campos diversos como son la visión por computador ([4], [5], [6]), la segmentación semántica ([7], [8], [9], [10]), la síntesis de series temporales ([11]), la edición de imagen ([12], [13], [14], [15]), el procesamiento del lenguaje natural ([16], [17], [18]), la generación de imagen a partir de texto ([19], [20], [21]) entre otros.

Para cualquier conjunto de datos, podemos hipotetizar que es posible definir una distribución de probabilidad $p_{data}$ representativa de la población representada por la muestra formada por el conjunto de datos. De ser esto posible, para cualquier valor de $x$ será posible establecer un valor $P_{data}(x)$ que determine la probabilidad de que $x$ pertenezca a la población. De existir una función de este tipo, sería una función discriminativa que dada una instancia permitiría conocer la probabilidad de pertenencia a la población. Los modelos generativos modelizan la distribución de probabilidad mencionada pero no proporcionan un valor de probabilidad, sino que generan instancias nuevas que pertenecen a distribuciones de probabilidad próximas a la que pretenden asemejar. Las GANs definen un esquema de aprendizaje que facilita la codificación de los atributos definitorios de la distribución de probabilidad en una red neuronal de manera que la red incorpore la información esencial que le permite generar instancias pertenecientes a distribuciones de probabilidad próximas a la que el conjunto de datos que pretende representar.

En la siguiente sección se presenta el esquema básico de la arquitectura GAN y su aspecto distintivo, la naturaleza de la función objetivo utilizada para su optimización. Posteriormente, se presentan las arquitecturas y funciones de objetivo derivadas, así como sus aplicaciones.

# 2 El concepto de redes antagónicas

La arquitectura GAN está formada por dos redes neuronales constituyentes: una denominada discriminadora ($D$) y otra generadora $G$. La red $G$ se encarga de generar nuevas instancias del mismo dominio que el del conjunto de datos de origen. La red $D$ se encarga de discriminar si los datos de entrada son reales, esto es pertenecientes al conjunto de datos de entrada o bien son ficticios, esto es generados artificialmente. Ambas redes se entrenan de manera conjunta de manera que $G$ maximice sus posibilidades de no ser detectada por $D$ y $D$ de forma que haga cada vez más sofisticados sus métodos de detección de los datos generados artificialmente por $G$. Estas dos redes adversarias compiten en un juego de suma cero en el que se hipotetiza que eventualmente llegan a un equilibrio de Nash [22].

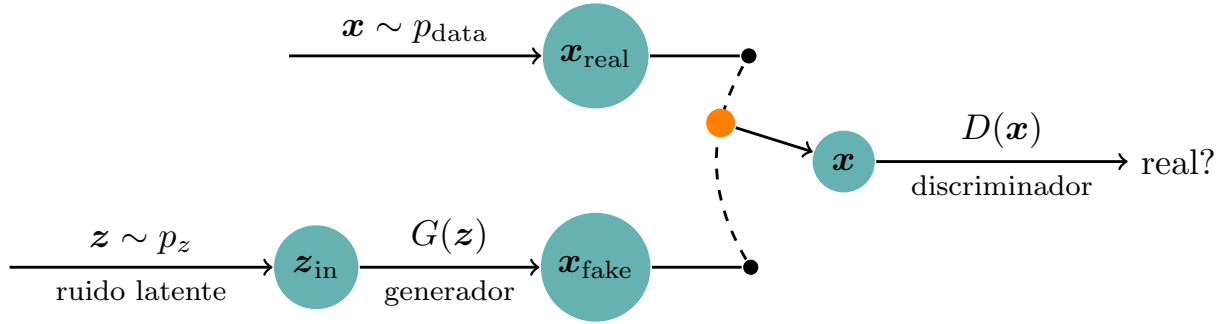

Figura 1: Diagrama representativo del proceso de entrenamiento de las redes adversarias generativas (GANs)

En la figura 2 se muestra un diagrama representativo del proceso de optimización de las GAN. Un vector $z$ es muestreo de una distribución de probabilidad aleatoria $p_z$, $z \sim p_z$ y alimentado como entrada a $G$. El propósito de la optimización es conseguir que $G(z) \sim p_g$ acabe siendo una estimación de la distribución de probabilidad $P_{data}$. Las GAN se optimizan la función min-max de un juego de suma cero expresado por la ecuación 1.

$$\min_G \max_D \mathbb{E}_{x \sim p_r} \log[D(\boldsymbol{x})] + \mathbb{E}_{\boldsymbol{z} \sim p_{\boldsymbol{z}}} \log[1 - D(G(\boldsymbol{z}))] \qquad (1)$$

Equivalentemente, sean $p_\theta$, $D_\omega$ las redes neuronales generadora y discriminadora de una GAN, siendo $\theta$ los parámetros de $G$ y $\omega$ los de $D$. Ambas redes se optimizan en conjunto con la función objetivo definida por la ecuación 2.

$$\min_\theta \max_\omega \mathbb{E}_{x \sim Q} \log[D_\omega(x)] + \mathbb{E}_{x \sim p_\theta} \log[1 - D_\omega(x))] \qquad (2)$$

Durante el proceso de optimización la red $D$ recibirá como entrada de manera aleatoria datos pertenecientes al conjunto de datos y otros procedentes de la red $G$. Se optimizará su funcionamiento para que su discriminación sea efectiva (ecuación 3).

$$\nabla_{\theta_D} \frac{1}{m} \sum_{i=1}^m [\log D(\boldsymbol{x}^{(i)}) + \log(1 - D(G(\boldsymbol{z}^{(i)})))] \qquad (3)$$

Al mismo tiempo, cuando $D$ reciba una entrada procedente de $G$, éste se optimizará para mejorar sus predicciones y hacer cada vez más difícil el papel de $D$. Esto únicamente se puede conseguir mejorando la calidad de los datos generados y haciéndolos más parecidos al conjunto de datos original (ecuación 4).

$$\nabla_{\theta_G} \frac{1}{m} \sum_{i=1}^m \log(1 - D(G(\boldsymbol{z}^{(i)}))) \qquad (4)$$

Se establece así una competición entre las dos redes (de ahí el nombre de adversarias) de forma que idealmente en el progreso de este proceso ambas mejoran su funcionamiento al punto que idealmente el generador acaba produciendo datos cada vez más parecidos a los del conjunto de datos original.

## 3 Ventajas e inconvenientes de las GAN

Desde su introducción en 2014, las GANs han despertado un gran interés sobre todo en el campo de la generación de imagen. Esto ha sido debido a que presentan una serie de ventajas sobre el otro paradigma dominante hasta el momento en lo que a modelos generativos se refiere, los VAEs [23]. Dichas ventajas son las siguientes:

- **Imágenes más nítidas**: Las GANs producen imágenes más nítidas que otros modelos generativos disponibles hasta el momento. Los modelos de difusión que veremos más adelante son una excepción posterior en este aspecto.
- **Tamaño configurable**: El tamaño de la variable aleatoria no está restringido pudiéndose enriquecer en caso de ser necesario.
- **Generador versátil**: El paradigma de diseño basado en GANs soporta distintos tipos de funciones generadoras, a diferencia de otros modelos generativos que pueden tener restricciones debido a su arquitectura. Los VAEs, por ejemplo, obligan a utilizar una función Gaussiana en la primera capa del Generador.

La arquitectura también tiene sus desventajas entre las que están las siguientes:

- **Colapso de modo**: Durante el entrenamiento sincronizado de generador y discriminador, el generador puede tener tendencia a reproducir únicamente un modo específico que es capaz de burlar al discriminador. A pesar de que este patrón puede estar minimizando la función objetivo, lo hace sin cubrir todo el dominio del conjunto de datos.
- **Desvanecimiento de gradientes**: A veces el discriminador se optimiza demasiado rápido en su función. En estos casos, los gradientes que propaga pueden ser demasiado bajos para asegurar la optimización del generador.
- **Inestabilidad**: A menudo durante el entrenamiento los parámetros de ambas redes fluctúan sin encontrar un punto de equilibrio. En estas circunstancias el generador tiene dificultades en encontrar un punto que genere imágenes de alta calidad.

## 4 Arquitecturas derivadas

En esta sección se presentan aquellas arquitecturas derivadas de la original que mejoran su rendimiento en alguna de las desventajas mencionadas.

### 4.1 GAN semi-supervisada (SGAN)

SGAN [24] incluye una variación en el discriminador que le permite aprovechar las ventajas de contar con datos supervisados. Consiste en añadir un cabezal adicional para la predicción de la clase de pertenencia. En aquellos casos reales que se conoce dicha clase, se utiliza el cabezal softmax de predicción para optimizar al discriminador. En aquellos que no se conozca, se utiliza la optimización vía clasificación binaria típica de la GAN convencional. Los resultados demuestran que este tipo de entrenamiento mejora las capacidades de SGAN respecto a la GAN original.

### 4.2 Conditional GAN (CGAN)

CGAN [25] modifica el método original introduciendo una entrada adicional tanto en el generador como en el discriminador. Esta entrada adicional sirve como condicionante para ambas funciones. Esta nueva información $y$ se fusiona en el generador con el muestreo de la variable aleatoria $z$ para posteriormente generar la nuevas instancias. Lo mismo ocurre en el discriminador donde $y$ se integra con los datos $x$ a analizar. La nueva función de optimización queda como indica la ecuación 5.

$$\min_G \max_D \mathbb{E}_{x \sim p_r} \log[D(\boldsymbol{x}|\boldsymbol{y})] + \mathbb{E}_{\boldsymbol{z} \sim p_{\boldsymbol{z}}} \log[1 - D(G(\boldsymbol{z}|\boldsymbol{y}))] \tag{5}$$

### 4.3 Red antagónica generativa convolucional profunda (DCGAN)

Las redes antagónicas generativas convolucionales profundas (DCGAN) fueron introducidas por primera vez en [26] como método para la generación de imágenes. Utilizan convoluciones en el discriminador y convoluciones traspuestas [27] en el generador. Además de mejoras en la resolución de las imágenes generadas, consiguen mejoras en la estabilidad del entrenamiento que en su estudio atribuyen a la introducción de las siguientes modificaciones:

- Sustitución de todas las capas de *pooling* de las dos redes. En el discriminador se utilizan núcleos con *stride* mayor que 1 y en el generador convoluciones traspuestas para aumentar el tamaño de la imagen.
- Uso de la normalización por lotes en las dos redes.
- En el discriminador se cambia la función de activación de ReLU a LeakyReLU [28]. En el generador se utilizan ReLU en todas las capas excepto en la última donde se usa la función de activación tangente hiperbólica (tanh).

### 4.4 Progressive GAN (PROGAN)

En [5] se introduce un método progresivo de entrenamiento y aumento de la resolución de las imágenes generadas que da muy buenos resultados. Muchas de las arquitecturas GAN más exitosas usan este método. En el trabajo citado se empieza entrenando una red generativa de 4x4 para ir añadiendo capas en el generador y discriminador conforme va avanzando el entrenamiento hasta conseguir resoluciones de salida de 1024x1024. Conforme se van añadiendo capas, todas las capas anteriores siguen estando sometidas a los cambios inherentes a la optimización de las redes.

### 4.5 Self-attention GAN (SAGAN)

Muchas de las implementaciones existentes hasta el momento fallan en la captura de patrones geométricos y estructurales de largo alcance. Se hipotetiza que la causa de esto es debida a la naturaleza convolucional de las arquitecturas generativas. Debido a ella, las dependencias son de corto alcance y requieren del paso a través de varias capas para ser resueltas. Existen diversas soluciones que se pueden aplicar para solucionar este aspecto. La primera sería aumentar el tamaño de las convoluciones con el aumento asociado de los requerimientos computacionales. Otra sería el incremento de la profundidad de las redes. Una tercera posibilidad, que es la propuesta por SAGAN [29] es la utilización de mecanismos de auto-atención en alguna de las capas de la red convolucional.

Los atributos de salida $\boldsymbol{x} \in \mathbb{R}^{C \times N}$ de una capa de la red neuronal se transforman en dos espacios $\boldsymbol{f}(\boldsymbol{x}) = \boldsymbol{W}_f \boldsymbol{x}$ y $\boldsymbol{g}(\boldsymbol{x}) = \boldsymbol{W}_g \boldsymbol{x}$ para posteriormente calcular la atención como indica la ecuación 6.

$$\beta_{j,i} = \frac{\exp(s_{i,j})}{\sum_{i=1}^{N} \exp(s_{i,j})} \quad \text{donde} \quad s_{i,j} = \boldsymbol{f}(\boldsymbol{x}_i)^T \boldsymbol{g}(\boldsymbol{x}_j) \tag{6}$$

$\beta_{j,i}$ indica la atención que está prestando a la región $i$ cuando está generando la $j$. $C$ es el número de canales y $N$ el número de atributos de la anterior capa. La salida de la capa de atención $\boldsymbol{o} = (\boldsymbol{o}_1, \boldsymbol{o}_2, ... \boldsymbol{o}_N) \in \mathbb{R}^{C \times N}$ se puede expresar como indica la ecuación 7.

$$\boldsymbol{o}_j = \boldsymbol{v}\Big(\sum_{i=1}^{N} \beta_{j,i} \boldsymbol{h}(\boldsymbol{x}_i)\Big), \, \boldsymbol{h}(\boldsymbol{x}_i) = \boldsymbol{W}_h \boldsymbol{x}_i, \, \boldsymbol{v}(\boldsymbol{x}_i) = \boldsymbol{W}_v \boldsymbol{x}_i \tag{7}$$

Donde $\boldsymbol{W}_g \in \mathbb{R}^{\bar{C} \times C}$, $\boldsymbol{W}_f \in \mathbb{R}^{\bar{C} \times C}$, $\boldsymbol{W}_h \in \mathbb{R}^{\bar{C} \times C}$ y $\boldsymbol{W}_v \in \mathbb{R}^{C \times \bar{C}}$ son matrices optimizadas en tiempo de entrenamiento, implementadas como convoluciones 1x1. En el artículo se fija $\bar{C} = C/8$ por ser más eficiente de cara a la computación y, según se indica, no afectar significativamente a los resultados.

El valor de auto-atención considerado se escala y se suma al valor del atributo de entrada, $\boldsymbol{y}_i = \gamma \boldsymbol{o}_i + \boldsymbol{x}_i$ siendo $\gamma$ un parámetro optimizable que se inicializa a cero. Esto tiene su lógica pues al inicio del entrenamiento se puede esperar resolver las dependencias locales para que una vez avanzada la optimización se afine resolviendo las dependencias de mayor alcance. Este mecanismo de atención se aplica tanto al generador como al discriminador. Ambos son optimizados minimizando una versión modificada de la función de optimización original, que toma la forma de la ecuación 8. Esta ecuación es la versión de máximo-margen (Hinge Loss) típica para la optimización de SVMs [30].

$$\begin{aligned} L_D &= -\mathbb{E}_{(x,y) \sim P_{data}}[\min(0, -1 + D(x, y))] - \mathbb{E}_{z \sim p_z, y \sim P_{data}}[\min(0, -1 - D(G(z), y))] \\ L_G &= -\mathbb{E}_{z \sim p_z, y \sim P_{data}} G(G(z), y) \end{aligned} \tag{8}$$

Junto con su propuesta presentan también una metodología de entrenamiento para estabilizar el inherentemente inestable proceso de optimización de las GANs.

### 4.6 BigGAN

BigGAN [31] es un tipo de GAN diseñada para la generación, mediante escalado, de imágenes de alta resolución. Incluye una serie de cambios incrementales respecto a las redes anteriormente mencionadas así como también algunas innovaciones.

Entre las mejoras incrementales de relevancia encontramos las siguientes:

1. Propuesta de arquitectura basada en SAGAN utilizando normalización espectral [32] tanto para $D$ como para $G$ y utilizando TTUR [33].
2. Al igual que SAGAN utiliza la función de pérdida Hinge como objetivo de optimización
3. Utiliza normalización por lotes condicionada a la clase (CBN) [34] (mediante proyección lineal) para proveer de información de la clase a $G$.
4. Utiliza un discriminador por proyección [35] para incorporar la información de la clase.

En cuanto a las innovaciones resaltar:

1. Incremento del tamaño de los lotes.
2. Incremento del tamaño de capa.
3. Adición de conexiones directas entre la variable latente $z$ y las capas intermedias de la red.
4. Uso de una variante de regularización ortogonal [36].

BigGAN consigue una mejora sustancial de la calidad de las imágenes generadas para tamaños de 128x128, 256x256 y 512x512, aumentando el número de parámetros (x4) y incrementando el tamaño del batch de entrenamiento (x8) respecto a SAGAN. Además de las modificaciones indicadas, introducen algunos cambios sobre las variables latentes, durante el proceso de inferencia consistentes en truncar los valores fuera de un rango determinado.

### 4.7 StyleGAN

StyleGAN [37] propone una nueva arquitectura para el generador, manteniendo el mismo diseño para el discriminador. En la figura 4.7 se muestra un esquema de la arquitectura propuesta para la red generadora.

StyleGAN tiene tres componentes característicos:

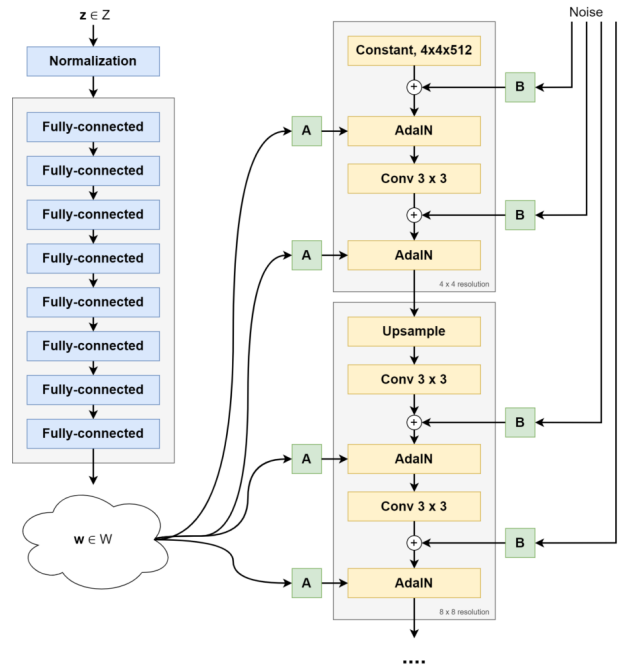

Figura 2: Diagrama de la arquitectura StyleGAN. Fuente: `https://github.com/christianversloot/machine-learning-articles/blob/main/stylegan-a-step-by-step-introduction.md`

1. Crecimiento progresivo del tamaño de la imagen generada al estilo de PROGAN (bloques en amarillo) $(4 \times 4 \to 8 \times 8 \to 16 \times 16 \to 32 \times 32 \to 64 \times 64 \to 128 \times 128 \to 256 \times 256 \to 512 \times 512 \to 1024 \times 1024)$

2. En los bloques de Upsample utiliza muestreo bilinear en vez de la copia del valor de los vecinos cercanos.

3. Sustitución del vector de entrada $z \sim P_z$ por una matriz de entrada constante de $4 \times 4 \times 512$ y se introduce una red de ruido gaussiano, que alimenta independientemente a cada una de las capas intermedias.

4. Introducción de la denominada *Red de mapeo del ruido*, que al paso por varias capas completamente conectadas, transforma una entrada aleatoria en una representación interna de los estilos. La salida de esta red actúa como entrada en los bloques AdaIN de cada capa fijando los parámetros de sesgo y escalado del bloque con la finalidad de actuar como estilos adaptativos.

5. AdaIN. Normalización adaptativa de las instancias: Introducida inicialmente en [38]. StyleGAN la utiliza tanto como capa de normalización como para establecer el estilo a través de los parámetros de escalado y sesgo que son derivados de las representaciones internas aprendidas por la red de mapeo de ruido.

# 5 f-GAN una generalización de las GANs

En las secciones previas, hemos presentado a las GANs como una herramienta para generar datos, que se optimiza para producir resultados que se parezcan a los datos de entrenamiento. Se ha explicado cómo este proceso se lleva a cabo a través de la optimización de un modelo parametrizable en forma de red neuronal, midiendo la diferencia entre la distribución del modelo y la real durante el proceso de entrenamiento.

En este apartado, exploraremos cómo distintas formas de medir esta distancia pueden dar lugar a diferentes funciones objetivo y, por lo tanto, a distintas redes resultantes. Este concepto se ha formalizado en el artículo [39], que proporciona una generalización del concepto de distancia en GANs para diferentes formas de medir la distancia entre distribuciones de probabilidad.

## 5.1 Introducción

Un modelo probabilístico es una representación formal que describe un evento o fenómeno en términos de una distribución de probabilidad, en lugar de proporcionar una respuesta única como lo hacen los modelos deterministas.

Estos modelos se utilizan para modelar procesos estocásticos, donde los resultados no están determinados de antemano y pueden variar con el tiempo o en función de las condiciones.

Cuando los principios que rigen un fenómeno son desconocidos o es demasiado complejo describirlos de manera computacionalmente eficiente, una forma de aproximar la solución es modelar su distribución de probabilidad a partir de muestras recopiladas del entorno. De esta manera, se pueden hacer predicciones y tomar decisiones basadas en una comprensión probabilística del fenómeno en cuestión.

### 5.2 Estimación de modelos probabilísticos

Suponiendo que existe una distribución de probabilidad real $Q$, se puede utilizar un modelo paramétrico $P$ para aproximar a $Q$. Para evaluar la validez de nuestro modelo debemos tener un medio de evaluar las diferencias entre ambas distribuciones. Sabemos que ambas variables son estocásticas, esto es, no están definidas por un único valor comparable sino que son distribuciones (funciones). Debemos disponer, por tanto, de una medida que nos permita comparar funciones. Una forma de hacer esto es realizar una abstracción del concepto de distancia para generalizarlo, no únicamente para medir distancias entre puntos, sino también para identificar distancias entre distribuciones de probabilidad. Para facilitar el proceso, puede ser necesario hacer algún tipo de suposición sobre $P$, como que su muestreo sea manejable, de cara a asegurar que las muestras del modelo sean comparables con las muestras reales; que $P$ tenga un gradiente manejable con respecto al muestreo de cara a poder utilizar técnicas de optimización y, finalmente, que tenga una función de probabilidad manejable de forma que pueda calcularse la probabilidad punto a punto.

### 5.3 GANs como modelos probabilísticos

Como ya sabemos, las GANs usan una combinación de dos redes neuronales para crear un modelo generativo con el objetivo de aprender a imitar una distribución real. El generador, como su nombre indica, genera una muestra modelo a partir de un vector aleatorio mediante el uso de una transformación determinista. El discriminador recibe muestras pertenecientes a la distribución real escogidas al azar y del modelo de generación y se entrena para diferenciarlas. Ambas redes están entrenadas de manera adversaria, una para generar imágenes lo más cercanas posible a la distribución real y la otra para detectar las muestras procedentes del generador. Aunque las GAN no proporcionan un valor de probabilidad de la imagen generada, es de esperar que este valor, aunque no sea conocido, exista.

### 5.4 Distancia entre distribuciones de probabilidad

Existen diferentes enfoques para definir la distancia entre las distribuciones de probabilidad. Podemos diferenciar principalmente entre tres enfoques diferentes:

1. Métricas de probabilidad integral
2. Reglas para puntuación
3. f-divergencias

#### 5.4.1 Métricas de probabilidad integral

Abordado por primera vez en las publicación [40], el punto clave de este método es que ambas distribuciones aparecen en forma de expectativa, lo que permite aproximarse a dicho valor mediante muestreo. La distancia de Wasserstein se deriva de estos métodos, y luego se usó con éxito como una distancia en Wasserstein-GAN [41].

#### 5.4.2 Reglas para puntuación

En [42] se propuso el enfoque de utilizar una puntuación para evaluar la adecuación de una distribución a otra. El punto es la optimización de la puntuación. La puntuación máxima se logra cuando ambas distribuciones son iguales.

#### 5.4.3 f-divergencias

Abordado por primera vez en [43], las distribuciones $P$ y $Q$ se requieren en forma de función de densidad. Las f-divergencias son una generalización de la divergencia Kullback-Leiber [44]. La expectativa de la distribución $Q$ se multiplica por una función convexa de la relación de verosimilitud entre $P$ y $Q$. Por lo general, no se aplica directamente porque normalmente no se dispone de la distribución de los datos de forma explícita.

En [45], [46] y en [1] se desarrollaron métodos para usar f-divergencias en problemas donde $P$ tiene forma de distribución y $Q$ tiene forma de expectativa y también donde ambas distribuciones tienen forma de expectativa. Esto permitió el uso de tales métricas en problemas donde solo se encuentran disponibles muestras de ambas distribuciones.

Matemáticamente, podemos definir la f-divergencia, denotada como $D_f(P \parallel Q)$, como la distancia entre dos distribuciones de probabilidad $P$ y $Q$ que se puede calcular mediante la ecuación 9.

$$D_f(Q \parallel P) = \int_{\mathscr{X}} p(\boldsymbol{x}) f\left(\frac{q(x)}{p(\boldsymbol{x})}\right) dx \tag{9}$$

donde $f$ es una función convexa, denominada función generador (nada que ver con el generador de las GANs), que cumple la propiedad $f(1) = 0$, esto es, que en aquellos puntos donde las dos funciones son iguales la distancia es cero.

La distancia siempre es ser mayor que 0 y es únicamente igual a cero cuando las dos distribuciones coinciden en todo el dominio $\mathscr{X}$.

| Nombre | $D_f(P \parallel Q)$ | Generador $f(u)$ |
|---|---|---|
| Total variation | $\frac{1}{2} \int \mid p(\boldsymbol{x}) - q(x) \mid dx$ | $\frac{1}{2} \mid u - 1 \mid$ |
| Kullback-Leibler | $\int p(\boldsymbol{x}) \log \frac{p(\boldsymbol{x})}{q(x)} dx$ | $u \log u$ |
| Reverse Kullback-Leibler | $\int q(x) \log \frac{q(x)}{p(\boldsymbol{x})} dx$ | $-\log u$ |
| Pearson $\chi^2$ | $\int \frac{(q(x) - p(\boldsymbol{x}))^2}{p(\boldsymbol{x})} dx$ | $(u-1)^2$ |
| Neyman $\chi^2$ | $\int \frac{(p(\boldsymbol{x}) - q(x))^2}{q(x)} dx$ | $\frac{(1-u)^2}{u}$ |
| Squared Hellinger | $\int (\sqrt{p(\boldsymbol{x})} - \sqrt{q(x)})^2 dx$ | $(\sqrt{u} - 1)^2$ |
| Jeffrey | $\int (p(\boldsymbol{x}) - q(x)) \log \left(\frac{p(\boldsymbol{x})}{q(x)}\right) dx$ | $(u-1)\log u$ |
| Jensen-Shannon | $\frac{1}{2} \int p(\boldsymbol{x}) \log \frac{2p(\boldsymbol{x})}{p(\boldsymbol{x})+q(x)} + q(x) \log \frac{2q(x)}{p(\boldsymbol{x})+q(x)} dx$ | $-(u+1)\log \frac{1+u}{2} + u \log u$ |
| Jensen-Shannon weighted | $\int p(\boldsymbol{x})\pi \log \frac{p(\boldsymbol{x})}{\pi p(\boldsymbol{x}) + (1-\pi)q(x)} + (1-\pi)q(x) \log \frac{q(x)}{\pi p(\boldsymbol{x}) + (1-\pi)q(x)} dx$ | $\pi u \log u + (1 - \pi + \pi u)\log(1 - \pi + \pi u)$ |
| GAN | $\int p(\boldsymbol{x}) \log \frac{2p(\boldsymbol{x})}{p(\boldsymbol{x})+q(x)} + q(x) \log \frac{2q(x)}{p(\boldsymbol{x})+q(x)} dx - \log 4$ | $u \log u - (u+1)\log(u+1)$ |
| $\alpha$-divergence | $\frac{1}{\alpha(\alpha-1)} \int \left( p(\boldsymbol{x}) \left[ \left(\frac{q(x)}{p(\boldsymbol{x})}\right)^{\alpha} - 1 \right] - \alpha(q(x) - p(\boldsymbol{x})) \right) dx$ | $\frac{1}{\alpha(\alpha-1)}(u^{\alpha} - 1 - \alpha(u-1))$ |

Cuadro 1: f-divergencias y generador asociado. Formas distintas de medir la distancia entre dos distribuciones de probabilidad $P$ y $Q$

### 5.4.4 Estimación de las f-divergencias mediante muestreo

En [45] se deriva un método variacional general para estimar las f-divergencias a partir del muestreo de las distribuciones $P$ y $Q$.

Del análisis de funciones convexas sabemos que toda función convexa $f$ tiene un conjugado de Fenchel $f^*$ para el que se cumple que $f(u) = \sup_{t \in dom_{f^*}} \{tu - f^*(t)\}$, esto es, que cualquier función convexa puede ser representada como un conjunto de máximos puntuales de funciones lineales. Siendo $f^*(t)$ el punto de intercepción con el eje de $y$ para cada punto $t$.

$$D_f(Q \parallel P) = \int_{\mathscr{X}} p(\boldsymbol{x}) f(\frac{q(x)}{p(\boldsymbol{x})}) dx \geq$$

$$\geq \sup_{T \in \mathscr{T}} \left( \int q(x) T(x) dx - \int_{\mathscr{X}} p(\boldsymbol{x}) f^*(T(x)) dx \right) =$$

$$= \sup_{T \in \mathscr{T}} \left( \mathbb{E}_{x \sim Q}[T(x)] - \mathbb{E}_{x \sim P}[f^*(T(x))] \right) \tag{10}$$

Las expresiones pueden ser convertidas a expectativas y utilizadas en un algoritmo de muestreo. Un punto a tener en cuenta es que para algunas divergencias, $f^*$ únicamente está definido para un dominio restringido. En esos casos, antes de utilizar la función objetivo, debe ser mapeada al dominio donde está definida.

### 5.4.5 Función variacional asociada al discriminador

En el artículo introductorio de las GAN [1] se demuestra que la optimización propuesta para la función de pérdida (eq. 2) es equivalente a minimizar la divergencia de Jensen-Shanon.

Comparando la expresión de la ecuación 10 con la función objetivo de GAN vemos que $T(x) = log(D_\omega(x))$. Confirmamos pues que la GAN original minimiza la divergencia Jensen-Shanon, una caso particular de la f-GAN.

## 5.5 Conclusión

En esta sección se ha generalizado el concepto de medida de la distancia entre distribuciones de probabilidad, utilizado en la función de pérdida de las GAN, para poderlo usar en un contexto más general con cualquier f-divergencia para la medida de la distancia entre la distribución de probabilidad del modelo y la original. En este contexto más general, se ha demostrado que la GAN original es un caso particular de optimización que utiliza la f-divergencia de Jensen-Shanon para la medida de la distancia entre distribuciones. Cualquiera de las medidas alternativas presentadas da lugar a optimizaciones equivalentes, a pesar de ser distintas, desde un punto de vista representativo.

## 5.6 Funciones de optimización mejoradas

En la anterior sección hemos introducido el concepto de f-divergencia para generalizar el concepto de distancia entre distribuciones y hemos nombrado otros métodos alternativos como las métricas de probabilidad integral. En esta sección presentaremos las opciones de optimización que en la práctica han demostrado ser más efectivas para mejorar los problemas relacionados con la optimización min-max original, esto es, el colapso de modo y el desvanecimiento de gradientes. Los objetivos presentados a parte de remediar estos aspectos también mejora la calidad de las imágenes generadas.

### 5.6.1 Wasserstein GAN (WGAN)

WGAN [41] resuelve el problema del desvanecimiento de gradientes y colapso de modo reemplazando la optimización de la f-divergencia de Jensen Shanon mediante el uso de la métrica de probabilidad integral EM (Earth mover distance [47]) también llamada distancia de Wasserstein. La ecuación 11 muestra la ecuación que la define.

$$W(p_r, p_g) = \inf_{\gamma \in \prod(p_r, p_g)} \mathbb{E}_{(x,y) \sim \gamma}[||x - y||] \tag{11}$$

donde $\prod(p_r, p_g)$ representa en conjunto de todas las distribuciones conjuntas $\gamma(x, y)$ cuyos marginales son $p_r$ y $p_g$. La distancia EM representa el mínimo coste necesario para transportar la "masa" de $p_r$ a $p_g$ para conseguir hacerlas iguales.

Las f-divergencias como KL y JS muestran problemas de inestabilidad cuando $p_r$ y $p_g$ están muy alejadas una de otra. EM es estable también en estos casos, además de ser continua, hecho que facilita la derivación de gradientes útiles para llevar a cabo la optimización. Un inconveniente es que el ínfimo de la ecuación 11 es intratable. Por esa razón los autores de WGAN estiman el coste de EM con la ecuación 12.

$$W(p_r, p_g) \approx \max_{w \sim \mathcal{W}} \mathbb{E}_{\boldsymbol{x} \sim p_r}[f_w(\boldsymbol{x})] - \mathbb{E}_{\boldsymbol{z} \sim p_z}[f_w(G(\boldsymbol{z}))] \tag{12}$$

donde $f_{w\, w \in \mathcal{W}}$ es una familia de funciones paramétricas que son K-Lipschitz para algún $K(||f||_L) \leq K)$.

Los autores proponen encontrar la mejor función $f_w$ que maximiza la eq. 12 propagando el gradiente $\mathbb{E}_{\boldsymbol{z} \sim p_z}[f_w(G(\boldsymbol{z}))]$, donde $g_\theta$ es el generador $g$ con parámetros $\theta$. $f_w$ puede ser representada por $D$ pero condicionada a ser K-Lipschitz. $w$ en $f_w$ representa a los parámetros de $D$ y el objetivo de $D$ es maximizar 12, que aproxima la distancia EM. Cuando $D$ es óptimo, 12 se aproxima a la distancia EM real y $G$ se optimiza para minimizar la ecuación 13.

$$-\min_{G} \mathbb{E}_{\boldsymbol{z} \sim p_z}[f_w(G(\boldsymbol{z}))] \tag{13}$$

WGAN suele tener unos gradientes más suaves y medibles en todo el dominio que otras f-divergencias y aprende mejor incluso cuando no está produciendo aún buenas imágenes.

### 5.6.2 GAN auto-supervisada (SSGAN)

SSGAN [48] utiliza un sistema similar a la CGAN [25] pero sin la necesidad de contar con etiquetado explícito. Además introduce un nuevo elemento en la función de pérdida, el objetivo del cual es prever el valor de una clase que se deriva directamente de la naturaleza de la imagen analizada. Concretamente, los autores predicen la rotación de la imagen de 4 posibles valores. Este valor es conocido y calculable sin necesidad de etiquetar las imágenes y se demuestra útil para mejorar las capacidades predictivas del discriminador, así como para aprender representaciones útiles para la generación. La nueva función de pérdida tiene la forma indicada en la ecuación 14.

$$L_G = -V(G,D) - \alpha \mathbb{E}_{\boldsymbol{x}\sim p_G}\mathbb{E}_{r\sim\mathbb{R}}[\log Q_D(R=r \mid \boldsymbol{x}^r)]$$
$$L_D = -V(G,D) - \beta \mathbb{E}_{\boldsymbol{x}\sim p_{data}}\mathbb{E}_{r\sim\mathbb{R}}[\log Q_D(R=r \mid \boldsymbol{x}^r)] \quad (14)$$

donde V(G,D) es el objetivo general de la GAN original (ecuación 1), $P_{data}$ y $P_G$ son las distribuciones real y del generador respectivamente, $r \in \mathbb{R}$ es la rotación seleccionada entre los ángulos permitidos ($R = \{0, 90, 180, 270\}$). Una imagen rotada $r$ grados se denota como $\boldsymbol{x}^r$ y $Q(R \mid \boldsymbol{x}^r)$ es la distribución predicha sobre los ángulos de rotación de la muestra. Esta función de pérdida fuerza a aprender representaciones internas de la rotación de forma supervisada hipotetizando que dichas representaciones serán útiles también para la generalización de las capacidades generativas de la GAN. SSGAN obtiene resultados comparables a los de CGAN sin la necesidad de etiquetado externo.

### 5.6.3 Normalización espectral (SNGAN)

En SNGAN [32] se propone el uso de normalización de los parámetros como medio de estabilización del entrenamiento del discriminador. Es una técnica que es computacionalmente eficiente y que puede ser aplicada de manera sencilla. Sabemos que si $D$ es una función k-Lipshitz, esto es, intuitivamente, que es una función continua sin variaciones abruptas, esto puede servir para estabilizar el entrenamiento de las GAN. SNGAN controla el valor de la constante de Lipschitz limitando la norma espectral de cada capa, normalizando la matriz de pesos de cada capa $W$ para satisfacer la restricción $\sigma(W) = 1$ (esto es, que el mayor valor singular de la matriz de pesos sea 1). Esto se consigue simplemente normalizando cada capa mediante la ecuación 15.

$$\bar{\boldsymbol{W}}_{SN}(\boldsymbol{W}) = \frac{\boldsymbol{W}}{\sigma(\boldsymbol{W})} \quad (15)$$

donde $\boldsymbol{W}$ representa a la matriz de pesos de cada capa. En el artículo citado se demuestra que esta operación hace que la constante de Lipschitz en el discriminador quede limitada a un máximo de 1, hecho que facilita la optimización.

SNGAN consigue mejoras importantes de los resultados comparado con las técnicas previas de estabilización del entrenamiento publicadas, entre las que se incluyen el recorte del valor de los pesos [41], la penalización de gradientes [49], [50], normalización batch [51], normalización de pesos [52], normalización de capas [53] y regularización ortonormal [36].

### 5.6.4 SphereGAN

SphereGAN [54] es una GAN que utiliza métrica de probabilidad integral (IPM) que usa una hiperesfera para ligar la IPM a la función objetivo y de esta forma mejorar la estabilidad del entrenamiento. La función objetivo utilizada es la indicada en la ecuación 16.

$$\min_G \max_D \sum_r \mathbb{E}_x[d_s^r(\boldsymbol{N}, D(\boldsymbol{x}))] - \sum_r \mathbb{E}_z[d_s^r(\boldsymbol{N}, D(G(\boldsymbol{z})))] \quad (16)$$

para $r = 1, ..., R$ donde $d_s^r$ mide la distancia con momento $r$ entre cada muestra y el polo norte de la hiperesfera, $\boldsymbol{N}$. El subíndice $s$ indica que $d_s^r$ está definido en $\mathbb{S}^n$.

Esta propuesta mejora a las anteriores funciones objetivo basadas en la distancia Wasserstein gracias a la definición de los IPM en la hiperesfera. Esto le permite prescindir de muchas de las restricciones que deben ser impuestas a $D$ para asegurar un entrenamiento estable en las anteriores propuestas, como las restricciones sobre la naturaleza Lischitz de las funciones requeridas por la distancia Wasserstein.

El funcionamiento de SphereGAN se podría describir de la forma siguiente: la red $G$ genera datos a partir de un vector aleatorio $z$. Imágenes reales y otras las procedentes de $G$ se alimentan a la red discriminadora $D$ que, a diferencia de las propuestas anteriores, da como resultado un vector de dimensión $n$. SphereGAN re-mapea este vector en una hiperesfera de n dimensiones utilizando transformaciones geométricas. Los puntos mapeados se utilizan para calcular los momentos geométricos centrados en el polo norte ($N$) de dicha hiper-esfera. La red discriminadora intenta maximizar las diferencias de momento entre las imágenes reales y las falsas, mientras que la generadora intenta conseguir lo contrario.

## 6 Una aplicación de las GAN: aumento de datos

El aumento de datos es una técnica habitual para mejorar el entrenamiento de los modelos en aquellas circunstancias donde se dispone de pocos datos. Podríamos decir que las circunstancias más habituales que requieren aumento de datos serían las siguientes:

- **Etiquetado limitado**: Disponer de pocos datos etiquetados.
- **Diversidad limitada**: Cuando el conjunto de entrenamiento falta variedad en los datos.
- **Datos restringidos**: Alguna de la información es sensible y no puede utilizarse directamente.

Los dos primeros casos se pueden resolver manualmente, poniendo recursos para etiquetar más datos, pero puede suponer un coste elevado. Una alternativa para el primer caso consiste en utilizar GANs para aumentar la cantidad de datos. SGAN, por ejemplo, permite generar nuevos datos anotados. El segundo caso es más habitual, pues suele ser habitual disponer de conjuntos de datos con clases desbalanceadas pobres en las clases raras.

## 7 Conclusiones

En este capítulo hemos introducido a las GAN como método para la generación de datos. Es un tipo de arquitectura especialmente estudiada en el campo de generación de imagen. Es un tipo de arquitectura con mucho potencial que tiene algunos problemas sobre todo relacionados con la estabilidad en el entrenamiento. En este capítulo hemos estudiado el diseño original, sus ventajas e inconvenientes así como algunas modificaciones destinadas a solventar los problemas de la propuesta original. Si bien hemos intentado describir los puntos que consideramos clave, el campo en cuestión es muy amplio, tanto en arquitecturas, como en aplicaciones. Este capítulo debe servir como punto de partida para complementarlo con la bibliografía citada así como con las futuras aportaciones.

# Referencias

[1] Ian Goodfellow, Jean Pouget-Abadie, Mehdi Mirza, Bing Xu, David Warde-Farley, Sherjil Ozair, Aaron Courville, and Yoshua Bengio. Generative adversarial networks. *Communications of the ACM*, 63(11):139–144, 2020.

[2] Antonia Creswell, Tom White, Vincent Dumoulin, Kai Arulkumaran, Biswa Sengupta, and Anil A Bharath. Generative adversarial networks: An overview. *IEEE signal processing magazine*, 35(1):53–65, 2018.

[3] Gilad Cohen and Raja Giryes. Generative adversarial networks. *arXiv preprint arXiv:2203.00667*, 2022.

[4] Gintare Karolina Dziugaite, Daniel M Roy, and Zoubin Ghahramani. Training generative neural networks via maximum mean discrepancy optimization. *arXiv preprint arXiv:1505.03906*, 2015.

[5] Tero Karras, Timo Aila, Samuli Laine, and Jaakko Lehtinen. Progressive growing of gans for improved quality, stability, and variation. *arXiv preprint arXiv:1710.10196*, 2017.

[6] Christian Ledig, Lucas Theis, Ferenc Huszár, Jose Caballero, Andrew Cunningham, Alejandro Acosta, Andrew Aitken, Alykhan Tejani, Johannes Totz, Zehan Wang, et al. Photo-realistic single image super-resolution using a generative adversarial network. In *Proceedings of the IEEE conference on computer vision and pattern recognition*, pages 4681–4690, 2017.

[7] Pauline Luc, Camille Couprie, Soumith Chintala, and Jakob Verbeek. Semantic segmentation using adversarial networks. *arXiv preprint arXiv:1611.08408*, 2016.

[8] Phillip Isola, Jun-Yan Zhu, Tinghui Zhou, and Alexei A Efros. Image-to-image translation with conditional adversarial networks. In *Proceedings of the IEEE conference on computer vision and pattern recognition*, pages 1125–1134, 2017.

[9] Ting-Chun Wang, Ming-Yu Liu, Jun-Yan Zhu, Andrew Tao, Jan Kautz, and Bryan Catanzaro. High-resolution image synthesis and semantic manipulation with conditional gans. In *Proceedings of the IEEE conference on computer vision and pattern recognition*, pages 8798–8807, 2018.

[10] Judy Hoffman, Eric Tzeng, Taesung Park, Jun-Yan Zhu, Phillip Isola, Kate Saenko, Alexei Efros, and Trevor Darrell. Cycada: Cycle-consistent adversarial domain adaptation. In *International conference on machine learning*, pages 1989–1998. Pmlr, 2018.

[11] Kay Gregor Hartmann, Robin Tibor Schirrmeister, and Tonio Ball. Eeg-gan: Generative adversarial networks for electroencephalograhic (eeg) brain signals. *arXiv preprint arXiv:1806.01875*, 2018.

[12] Tamar Rott Shaham, Tali Dekel, and Tomer Michaeli. Singan: Learning a generative model from a single natural image. In *Proceedings of the IEEE/CVF International Conference on Computer Vision*, pages 4570–4580, 2019.

[13] Guillaume Lample, Neil Zeghidour, Nicolas Usunier, Antoine Bordes, Ludovic Denoyer, and Marc'Aurelio Ranzato. Fader networks: Manipulating images by sliding attributes. *Advances in neural information processing systems*, 30, 2017.

[14] Rameen Abdal, Peihao Zhu, Niloy J Mitra, and Peter Wonka. Styleflow: Attribute-conditioned exploration of stylegan-generated images using conditional continuous normalizing flows. *ACM Transactions on Graphics (ToG)*, 40(3):1–21, 2021.

[15] Weihao Xia, Yulun Zhang, Yujiu Yang, Jing-Hao Xue, Bolei Zhou, and Ming-Hsuan Yang. Gan inversion: A survey. *IEEE Transactions on Pattern Analysis and Machine Intelligence*, 2022.

[16] William Fedus, Ian Goodfellow, and Andrew M Dai. Maskgan: better text generation via filling in the_. *arXiv preprint arXiv:1801.07736*, 2018.

[17] Nikolay Jetchev, Urs Bergmann, and Roland Vollgraf. Texture synthesis with spatial generative adversarial networks. *arXiv preprint arXiv:1611.08207*, 2016.

[18] Jiaxian Guo, Sidi Lu, Han Cai, Weinan Zhang, Yong Yu, and Jun Wang. Long text generation via adversarial training with leaked information. In *Proceedings of the AAAI conference on artificial intelligence*, volume 32, 2018.

[19] Aditya Ramesh, Mikhail Pavlov, Gabriel Goh, Scott Gray, Chelsea Voss, Alec Radford, Mark Chen, and Ilya Sutskever. Zero-shot text-to-image generation. In *International Conference on Machine Learning*, pages 8821–8831. PMLR, 2021.

[20] Alec Radford, Jong Wook Kim, Chris Hallacy, Aditya Ramesh, Gabriel Goh, Sandhini Agarwal, Girish Sastry, Amanda Askell, Pamela Mishkin, Jack Clark, et al. Learning transferable visual models from natural language supervision. In *International Conference on Machine Learning*, pages 8748–8763. PMLR, 2021.

[21] Or Patashnik, Zongze Wu, Eli Shechtman, Daniel Cohen-Or, and Dani Lischinski. Styleclip: Text-driven manipulation of stylegan imagery. In *Proceedings of the IEEE/CVF International Conference on Computer Vision*, pages 2085–2094, 2021.

[22] Monireh Mohebbi Moghadam, Bahar Boroomand, Mohammad Jalali, Arman Zareian, Alireza DaeiJavad, and Mohammad Hossein Manshaei. Game of gans: Game theoretical models for generative adversarial networks. *arXiv preprint arXiv:2106.06976*, 2021.

[23] Diederik P Kingma and Max Welling. Auto-encoding variational bayes. *arXiv preprint arXiv:1312.6114*, 2013.

[24] Augustus Odena. Semi-supervised learning with generative adversarial networks. *arXiv preprint arXiv:1606.01583*, 2016.

[25] Mehdi Mirza and Simon Osindero. Conditional generative adversarial nets. *arXiv preprint arXiv:1411.1784*, 2014.

[26] Alec Radford, Luke Metz, and Soumith Chintala. Unsupervised representation learning with deep convolutional generative adversarial networks. *arXiv preprint arXiv:1511.06434*, 2015.

[27] Vincent Dumoulin and Francesco Visin. A guide to convolution arithmetic for deep learning. *arXiv preprint arXiv:1603.07285*, 2016.

[28] Andrew L Maas, Awni Y Hannun, Andrew Y Ng, et al. Rectifier nonlinearities improve neural network acoustic models. In *Proc. icml*, volume 30, page 3. Citeseer, 2013.

[29] Han Zhang, Ian Goodfellow, Dimitris Metaxas, and Augustus Odena. Self-attention generative adversarial networks. In *International conference on machine learning*, pages 7354–7363. PMLR, 2019.

[30] Koby Crammer and Yoram Singer. On the algorithmic implementation of multiclass kernel-based vector machines. *Journal of machine learning research*, 2(Dec):265–292, 2001.

[31] Andrew Brock, Jeff Donahue, and Karen Simonyan. Large scale gan training for high fidelity natural image synthesis. *arXiv preprint arXiv:1809.11096*, 2018.

[32] Takeru Miyato, Toshiki Kataoka, Masanori Koyama, and Yuichi Yoshida. Spectral normalization for generative adversarial networks. *arXiv preprint arXiv:1802.05957*, 2018.

[33] Martin Heusel, Hubert Ramsauer, Thomas Unterthiner, Bernhard Nessler, and Sepp Hochreiter. Gans trained by a two time-scale update rule converge to a local nash equilibrium. *Advances in neural information processing systems*, 30, 2017.

[34] Harm De Vries, Florian Strub, Jérémie Mary, Hugo Larochelle, Olivier Pietquin, and Aaron C Courville. Modulating early visual processing by language. *Advances in Neural Information Processing Systems*, 30, 2017.

[35] Takeru Miyato and Masanori Koyama. cgans with projection discriminator. *arXiv preprint arXiv:1802.05637*, 2018.

[36] Andrew Brock, Theodore Lim, James M Ritchie, and Nick Weston. Neural photo editing with introspective adversarial networks. *arXiv preprint arXiv:1609.07093*, 2016.

[37] Tero Karras, Samuli Laine, and Timo Aila. A style-based generator architecture for generative adversarial networks. In *Proceedings of the IEEE/CVF conference on computer vision and pattern recognition*, pages 4401–4410, 2019.

[38] Xun Huang and Serge Belongie. Arbitrary style transfer in real-time with adaptive instance normalization. In *Proceedings of the IEEE international conference on computer vision*, pages 1501–1510, 2017.

[39] Sebastian Nowozin, Botond Cseke, and Ryota Tomioka. f-gan: Training generative neural samplers using variational divergence minimization. *Advances in neural information processing systems*, 29, 2016.

[40] Bharath K Sriperumbudur, Arthur Gretton, Kenji Fukumizu, Bernhard Schölkopf, and Gert RG Lanckriet. Hilbert space embeddings and metrics on probability measures. *The Journal of Machine Learning Research*, 11:1517–1561, 2010.

[41] Martin Arjovsky, Soumith Chintala, and Léon Bottou. Wasserstein generative adversarial networks. In *International conference on machine learning*, pages 214–223. PMLR, 2017.

[42] Tilmann Gneiting and Adrian E Raftery. Strictly proper scoring rules, prediction, and estimation. *Journal of the American statistical Association*, 102(477):359–378, 2007.

[43] Syed Mumtaz Ali and Samuel D Silvey. A general class of coefficients of divergence of one distribution from another. *Journal of the Royal Statistical Society: Series B (Methodological)*, 28(1):131–142, 1966.

[44] Solomon Kullback and Richard A Leibler. On information and sufficiency. *The annals of mathematical statistics*, 22(1):79–86, 1951.

[45] XuanLong Nguyen, Martin J Wainwright, and Michael I Jordan. Estimating divergence functionals and the likelihood ratio by convex risk minimization. *IEEE Transactions on Information Theory*, 56(11):5847–5861, 2010.

[46] Mark Reid, Robert Williamson, et al. Information, divergence and risk for binary experiments. 2011.

[47] Frank L Hitchcock. The distribution of a product from several sources to numerous localities. *Journal of mathematics and physics*, 20(1-4):224–230, 1941.

[48] Ting Chen, Xiaohua Zhai, Marvin Ritter, Mario Lucic, and Neil Houlsby. Self-supervised gans via auxiliary rotation loss. In *Proceedings of the IEEE/CVF conference on computer vision and pattern recognition*, pages 12154–12163, 2019.

[49] Huikai Wu, Shuai Zheng, Junge Zhang, and Kaiqi Huang. Gp-gan: Towards realistic high-resolution image blending. In *Proceedings of the 27th ACM international conference on multimedia*, pages 2487–2495, 2019.

[50] Lars Mescheder, Andreas Geiger, and Sebastian Nowozin. Which training methods for gans do actually converge? In *International conference on machine learning*, pages 3481–3490. PMLR, 2018.

[51] Sergey Ioffe and Christian Szegedy. Batch normalization: Accelerating deep network training by reducing internal covariate shift. In *International conference on machine learning*, pages 448–456. PMLR, 2015.

[52] Tim Salimans and Durk P Kingma. Weight normalization: A simple reparameterization to accelerate training of deep neural networks. *Advances in neural information processing systems*, 29, 2016.

[53] Jimmy Lei Ba, Jamie Ryan Kiros, and Geoffrey E Hinton. Layer normalization. *ArXiv e-prints*, pages arXiv–1607, 2016.

[54] Sung Woo Park and Junseok Kwon. Spheregan: Sphere generative adversarial network based on geometric moment matching and its applications. *IEEE Transactions on Pattern Analysis and Machine Intelligence*, 2020.